\documentclass[times,twocolumn,final,authoryear]{elsarticle}

\usepackage{prlettersArxiv}
\usepackage{framed,multirow}

\usepackage{amssymb}
\usepackage{amsmath,graphicx,amsfonts}
\usepackage{subcaption}
\usepackage{booktabs}

\usepackage{tabularx,ragged2e}
\newcolumntype{C}{>{\Centering\arraybackslash}X}  
\newcolumntype{b}{>{\Centering}X}
\newcolumntype{s}{>{\Centering\hsize=.5\hsize}X}
\newcolumntype{j}{>{\Centering\hsize=.4\hsize}X}
\newcolumntype{k}{>{\Centering\hsize=.3\hsize}X}
\newcolumntype{y}{>{\Centering\hsize=.2\hsize}X}

\usepackage{url}
\usepackage{xcolor}
\definecolor{newcolor}{rgb}{.8,.349,.1}

\begin{document}

\thispagestyle{empty}

\clearpage
\thispagestyle{empty}
\ifpreprint
  \vspace*{-1pc}
\fi

\ifpreprint
  \setcounter{page}{1}
\else
  \setcounter{page}{1}
\fi

\begin{frontmatter}
\title{End-to-End Visual Speech Recognition for Small-Scale Datasets}

\author[1]{Stavros \snm{Petridis}\corref{cor1}} 
\cortext[cor1]{Corresponding author: 
  }
\ead{stavros.petridis04@imperial.ac.uk}
\author[1]{Yujiang \snm{Wang}}
\author[1]{Pingchuan \snm{Ma}}
\author[1]{Zuwei \snm{Li}}
\author[1]{Maja \snm{Pantic}}

\address[1]{Imperial College London, London, UK}

\received{1 May 2013}
\finalform{10 May 2013}
\accepted{13 May 2013}
\availableonline{15 May 2013}
\communicated{S. Sarkar}

\begin{abstract}
Visual speech recognition models traditionally consist of two stages, feature extraction and classification. Several deep learning approaches have been recently presented aiming to replace the feature extraction stage by automatically extracting features from mouth images. However, research on simultaneously learning features and performing classification remains limited. In addition, most of the existing methods require large amounts of data in order to achieve state-of-the-art performance, otherwise they under-perform.
In this work, an end-to-end visual speech recognition system is presented based on fully-connected layers and Long-Short Memory (LSTM) networks which is suitable for small-scale datasets. The model consists of two streams: one which extract features directly from the mouth images and one which extracts features from the difference images. A Bidirectional LSTM (BLSTM) is used for modelling the temporal dynamics in each stream which are then fused via another BLSTM. An absolute improvement of 0.6\%, 3.4\%, 3.9\%, 11.4\% over the state-of-the-art is reported on the OuluVS2, CUAVE, AVLetters and AVLetters2 databases, respectively. 
\end{abstract}


\end{frontmatter}


\section{Introduction}
\label{sec:intro}

Visual speech recognition or lip-reading is the process of recognising speech by observing only the lip movements, i.e., the audio signal is ignored. The first works in the field  \citep{Zhao2009,Potamianos2003,Dupont2000,matthews2002extraction} extract features from a mouth region of interest (ROI) and attempt to model their dynamics in order to recognise speech. Lip-reading systems can enable the use of silent interfaces and also enhance acoustic speech recognition in noisy environments since the visual signal is not affected by noise. 

Traditionally, two stages have been used for visual speech recognition systems: feature extraction from the mouth region of interest (ROI) and classification \citep{Potamianos2003, Dupont2000, Zhou2011}. 
Dimensionality reduction/compression methods, like Discrete Cosine Transform (DCT), are the most common feature extraction approach which results in a compact representation of the mouth ROI. In the second stage, the temporal evolution of the features is modelled by a dynamic classifier, like Hidden Markov Models (HMMs) or Long-Short Term Memory (LSTM) recurrent neural networks. 

\looseness - 1
Several deep learning approaches \citep{ninomiya2015integration,ngiam2011multimodal,petridis2016deep,sui2014extracting,chung2016lip} have been recently presented which automatically extract features from the pixels and replace the traditional feature extraction stage. Few end-to-end approaches have also been proposed which attempt to jointly learn the extracted features and perform visual speech classification \citep{petridis2017deepVisualSpeech,Chung17cvpr,wand2016lipreading,assael2016lipnet,stafylakis2017combining}. This has led to a new generation of deep-learning-based lipreading systems which significantly outperform the traditional approaches.

The vast majority of modern deep learning approaches require large amounts of data in order to achieve state-of-the-art performance and their success in smaller datasets has been modest. This has led to some researchers claiming that deep learning methods do not perform well on simple tasks and small-scale datasets. Hence, traditional visual speech recognition methods are a better choice when large datasets are not available \citep{fernandez2018survey}.

\looseness - 1
In this paper, an end-to-end visual speech recognition system is presented which  learns simultaneously the feature extraction and classification stages and is suitable for small-scale datasets where large deep models do not perform so well. The model is an improved version of the model presented in our previous work \citep{petridis2017deepVisualSpeech} and consists of two streams. One stream encodes static information and uses raw mouth ROIs as input. The other stream encodes local temporal dynamics and takes as input difference (diff) images. The temporal dynamics in each stream are modelled by a BLSTM and stream fusion takes place via another BLSTM. 

We perform experiments on four different datasets, OuluVS2, CUAVE, AVLetters and AVLetters2 which have  been used as the main lip-reading benchmarks before the introduction of very large lip-reading datasets and traditional lip-reading methods still achieve competetive results. A significant absolute improvement on the state-of-the-art classification rate is reported on all datasets.

\begin{figure}[t]

  \centering
\includegraphics[width=0.7\linewidth]{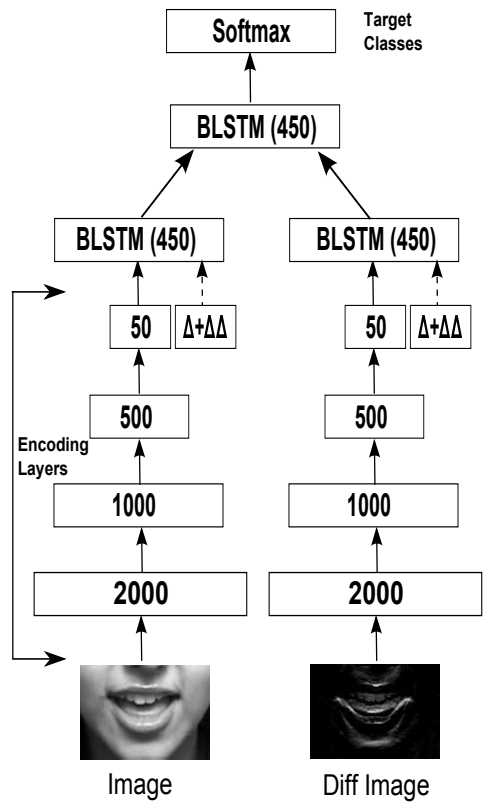}

\caption{Overview of the end-to-end visual speech recognition system. The model consists of two streams which extract features directly from the raw mouth ROI and the diff mouth ROI (in order to capture local temporal dynamics). The first and second derivatives ($\Delta$ and $\Delta\Delta$, respectively) features are also computed  and appended to the bottleneck layer. A BLSTM is used to model the temporal dynamics in each stream. The temporal dynamics across streams are modelled by another BLSTM. }
\label{fig:system}
\end{figure}

\section{Related Work}

In the first generation of deep models, deep bottleneck architectures \citep{ngiam2011multimodal,hu2016temporal,ninomiya2015integration,mroueh2015,takashima2016audio,petridis2016deep} were used to reduce the dimensionality of various visual and audio features extracted from the mouth ROIs and the audio signal. Then these features are fed to a classifier like a Support Vector Machine (SVM) or an HMM. Ngiam et al. \citep{ngiam2011multimodal} applied principal component analysis (PCA) to the mouth ROIs and bottleneck features were extracted with a deep autoencoder. Then the utterance features were fed to an SVM ignoring the temporal dynamics of the speech.  Ninomiya et al. \citep{ninomiya2015integration} followed a similar approach but the temporal dynamics were taken into account by an HMM. Another similar approach was  proposed by Sui et al. \citep{sui2014extracting} who extracted bottleneck features from local binary patterns which were concatenated with DCT features and fed to an HMM. Similar ideas have also been proposed for audiovisual speech recognition \citep{huang2013audio, mroueh2015, sui2015listening} where a shared representation of the input audio and visual features is extracted from the bottleneck layer.

In the second generation of deep models, deep bottleneck architectures were used which extract bottleneck features directly from the pixels. Li \citep{Li2016} extracted bottleneck features from dynamic representations of images with a convolutional neural network (CNN)  which were then fed to an HMM for classification.
In our previous work \citep{petridis2016deep}, bottleneck features were extracted directly from raw mouth ROIs by a deep feedforward network which were then fed to an LSTM network for classification.  Noda et al. \citep{noda2015audio} predicted the phoneme that corresponds to an input mouth ROI using a CNN, and then an HMM is used together with audio features in order to classify an utterance.

In the third generation of deep models, few end-to-end works have been presented which extract features directly from the mouth ROI pixels and perform classification. The main approaches followed can be divided into two groups. In the first one, fully connected layers are used to extract features and LSTM layers model the temporal dynamics of the sequence \citep{petridis2017deepVisualSpeech,wand2016lipreading}. In the second group, either 3D CNNs are used \citep{assael2016lipnet,shillingford2018large} or 3D convolutional layers followed by residual networks (ResNet) \citep{stafylakis2017combining} and then combined with LSTMs or Gated Recurrent Units (GRUs). 

These works have also been extended to audio-visual models. Chung et al. \citep{Chung17cvpr} applied an attention mechanism to both the mouth ROIs and MFCCs for continuous speech recognition. Petridis et al. \citep{end2endAV} used fully connected layers together with LSTMs  in order to extract features directly from raw images and spectrograms and perform classification on the OuluVS2 database \citep{Anina2015}. This method has been extended to extract features directly from raw images and audio waveforms using ResNets and bidirectional GRUs \citep{petridis2018AVspeech}.




%
%

\section{Databases}
\label{sec:Databases}

\looseness - 1
The databases used in this study are the OuluVS2 \citep{Anina2015}, AVLetters \citep{matthews2002extraction}, CUAVE \citep{Patterson:2002} and AVLetters2 \citep{cox2008challenge}. Fifty two speakers exist in the OuluVS2 database who repeat 3 times each of the 10 utterances, i.e., there are 156 examples per utterance. The following utterances are included in the dataset: ``Excuse me", ``Goodbye", ``Hello", ``How are you", ``Nice to meet you", ``See you", ``I am sorry", ``Thank you", ``Have a good time", ``You are welcome".  The provided mouth ROIs are used and they are downscaled to 26 by 44.

The AVLetters contains 10 speakers saying 3
times the letters A to Z, so in total there are 30 utterances per
letter. The mouth ROIs are provided and they are downscaled to 30 by 40.

\looseness - 1
The CUAVE dataset contains 36 subjects who repeat each digit (from 0 to 9)   5 times each, i.e, there are 180 examples per digit. The normal portion of the database is used which contains frontal facing speakers. The Dlib facial point tracker \citep{Kazemi_2014_CVPR} is used to track sixty eight points on the face. Then the faces are registered to a neutral reference frame in order to normalise them for rotation and size differences. An affine transform is used for this purpose using 5 stable points, two eyes corners in each eye and the tip of the nose. The center of the mouth is located based on the tracked mouth points 
and a bounding box
with size 90 by 150 is used to extract the mouth ROI which is then 
downscaled to 30 by 50.

The AVLetters2 contains 5 speakers saying 7
times the letters A to Z, so in total there are 35 utterances per
letter. The faces are first tracked and aligned using the same approach as in the CUAVE dataset. Then a bounding box, around the mouth centre, is extracted and downscaled to 30 by 45.

\section{End-To-End Visual Speech Recognition}
\label{sec:End2End}
The proposed deep learning model for visual speech recognition  consists of two independent streams, as shown in Fig.  \ref{fig:system}, which extract features directly from the raw input. Static information is mainly encoded  by the first stream which extracts features directly from the raw mouth ROI. Local temporal dynamics are modelled by the second stream which extracts features from the diff mouth ROI (computed by taking the difference between two consecutive frames). 

Both streams consist of two parts: an encoder and a BLSTM. The encoder follows a bottleneck architecture which compresses the high dimensional input image to a low dimensional representation.  It consists of 3 fully connected hidden layers of sizes 2000, 1000 and 500, respectively, with rectified linear units used as activation units similarly to \citep{hinton2006reducing}. This is followed by a linear bottleneck layer of size 50.  The first and second derivatives ($\Delta$ and $\Delta\Delta$ features, respectively) \citep{young2002htk}  are also computed, based on the bottleneck features, and they are appended to the bottleneck layer. In this way, the encoding layers are forced to learn compact representations which are not only discriminative for the task at hand but also produce discriminative $\Delta$ and $\Delta\Delta$ features. This is in contrast to the traditional approaches which have no control over the discriminative power of the $\Delta$ and $\Delta\Delta$ features which are pre-computed at the input level. 

The  BLSTM layer is added on top of the encoding layers in order to model the temporal dynamics of the features in each stream. The two streams are fused by concatenating the BLSTM outputs of each stream and feeding them to another BLSTM. A softmax layer is used as the output layer which provides a label for each input frame. The entire system is trained end-to-end so the feature extraction and classification layers are trained jointly. In other words, the encoding layers are trained to extract features from mouth ROI pixels which are useful for classification using BLSTMs.

\subsection{Single Stream Training}

\noindent
\textbf{Initialisation:} Each stream is first trained independently. Restricted Boltzmann Machines (RBMs) \citep{hinton2012practical} are used to pre-train in a greedy layer-wise manner the encoding layers. Four Gaussian RBMs  are used  since the input (pixels) is real-valued and the hidden layers are either rectified linear or linear (bottleneck layer). Each RBM is trained for 20 epochs using contrastive divergence with a mini-batch size of 100 and a fixed learning rate of 0.001. In addition, L2 regularisation is applied with a coefficient of 0.0002. 


\noindent \looseness - 1
\textbf{End-to-End Training:} A BLSTM is added on top of the pre-trained encoding layers and its weights are initialised using Glorot initialisation \citep{glorot2010understanding}.
Then the model is trained end-to-end using Adam with a mini-batch size of 10 utterances. A learning rate of 0.0003 was used since the default one of 0.001 led to unstable training. In order to avoid overfitting early stopping with a delay of 5 epochs was also used. In addition, gradient clipping was applied to the BLSTM layers. 

\subsection{Two-Stream Training}

\noindent
\textbf{Initialisation:} Each stream in the final model is initialised based on the corresponding single streams which have been already trained. Then  on top of all streams a BLSTM is added in order to fuse the outputs of the single streams. The BLSTM weights are initialised using Glorot initialisation.

\noindent
\textbf{End-to-End Training:} Finally, the two-stream model is fine-tuned using Adam with learning rate 0.0001. Similarly to single stream training early stopping and gradient clipping were also applied .

\begin{table}[t!]
\renewcommand{\tabcolsep}{7pt}
\caption{Classification Accuracy on the OuluVS2 database. The protocol suggested in \citep{ouluVS2}, where the training and validation sets consist of 40 subjects and the test set contains 12 subjects, is used for evaluating the end-to-end models. ``Mean (Std)'' refers to the mean classification accuracy over ten runs and the corresponding standard deviation, while ``Max'' reports the maximum classification accuracy. $^\ast$In cross-view training, the model is first trained with data from all views and then fine-tuned with data from the corresponding view. $^{\ast\ast}$These models are pretrained on the LRW dataset \citep{chung2016lip}, which is a large database, and then fine-tuned on OuluVS2.  DA: Data Augmentation, TDNN: Time-Delay Neural Network, LVM: Latent Variable Models}
\label{tab:resultsOuluVS}
\centering
\begin{tabularx}{\columnwidth}{bsy}
\toprule  Method & Mean (Std)  & Max \\
\midrule End-to-End (Raw Image)   &91.8 (1.1) &94.7  \\ 
\midrule End-to-End (Diff Image)   &90.3 (1.2) &92.2  \\ 
\midrule End-to-End (Raw + Diff Images)  & \textbf{93.6 (1.0)} & \textbf{95.6} \\
\midrule Multitask CNN + BLSTM \citep{han2017multi} & - & 95.0 \\
\midrule CNN pretrained on LRW dataset + DA + LSTM$^{\ast\ast}$ \citep{chung2016out} &-& 94.1  \\
\midrule CNN pretrained on LRW dataset + DA$^{\ast\ast}$ \citep{chung2016lip} &-& 93.2  \\
\midrule Autoencoder + TDNN + LSTM \citep{koumparoulis2018deep} & - & 90.0 \\
\midrule maxout-CNN-BLSTM \citep{fung2018end} & - & 87.6 \\
\midrule CNN + DA \citep{saitoh2016concatenated}  &- & 85.6   \\ 
\midrule CNN + LSTM, Cross-view Training$^\ast$ \citep{lee2016multi} &-& 82.8 \\
\midrule End-to-end CNN + LSTM \citep{lee2016multi} &-& 81.1  \\
\midrule DCT + HMM \citep{ouluVS2Results} &-& 74.8 \\
\midrule PCA Network + LSTM + GMM-HMM \citep{zimmermann2016visual} & -&74.1 \\
\midrule Raw Pixels + LVM  \citep{ouluVS2Results} &-& 73.0  \\
\bottomrule
\vspace{-0.5cm}
\end{tabularx} 
\end{table}

\begin{table}[t!]
\renewcommand{\tabcolsep}{7pt}
\caption{Classification Accuracy on the CUAVE database. The end-to-end models are evaluated using the protocol suggested in \citep{ngiam2011multimodal, Srivastava:2014} where 18 subjects are used for training and validation and 18 for testing. ``Mean (Std)'' refers to the mean classification accuracy over ten runs and the corresponding standard deviation, while ``Max'' reports the maximum classification accuracy. $\ddagger$ This model is trained and tested using a 9-fold cross validation. $\ast$This model is trained on 28 subjects and tested on 8 subjects. $\dagger$ These models are trained and tested using a 6-fold cross validation. }
\label{tab:resultsCUAVE}
\centering
\begin{tabularx}{\columnwidth}{bsy}
\toprule  Method & Mean (Std)  & Max \\
\midrule End-to-End (Raw Image)   &85.5 (0.7) &86.4  \\ 
\midrule End-to-End (Diff Image)   &82.8 (1.0) &83.9  \\ 
\midrule End-to-End (Raw + Diff Images)  & \textbf{87.3 (0.7)} & \textbf{88.4} \\
\midrule SVM + MKL \citep{benhaim2013designing} $\ddagger$ & - & 85.0 \\
\midrule Visemic AAM + HMM \citep{Papandreou2009} $\dagger$  &-& 83.0 \\
\midrule Patch-based Features + HMM \citep{Lucey:2006} $\ast$ &-& 77.1 \\
\midrule AAM +HMM \citep{Papandreou2007} $\dagger$  &-& 75.7 \\
\midrule Deep Boltzmann Machines + SVM \citep{Srivastava:2014} & 69.0 (1.5) & - \\
\midrule   Deep Autoencoder + SVM \citep{ngiam2011multimodal} & 68.7 (1.8) & - \\ 
\bottomrule
\vspace{-0.5cm}
\end{tabularx} 
\end{table}

\begin{table}[t!]
\renewcommand{\tabcolsep}{7pt}
\caption{Classification Accuracy on the AVLETTERS database. The end-to-end models are trained using the standard evaluation protocol \citep{matthews2002extraction} where the first 2 utterances of each subjects are used for training and the last one for testing. ``Mean (Std)'' refers to the mean classification accuracy over ten runs and the corresponding standard deviation, while ``Max'' reports the maximum classification accuracy. PLS: Partial Least Squares, DBNF: Deep BottleNeck Features, LBP-TOP: Local Binary Patterns-Three Orthogonal Planes.}
\label{tab:resultsAVLETTERS}
\centering
\begin{tabularx}{\columnwidth}{bsy}
\toprule  Method & Mean (Std)  & Max \\
\midrule End-to-End (Raw Image)   &65.9 (2.1) &68.9  \\ 
\midrule End-to-End (Diff Image)   &57.3 (1.8) &60.0  \\ 
\midrule End-to-End (Raw + Diff Images)  & \textbf{66.3 (2.0)} & \textbf{69.2} \\
\midrule Manifold Kernel PLS \citep{Bakry_2013_CVPR} &-& 65.3 \\
\midrule   Deep Boltzmann Machines + SVM \citep{Srivastava:2014} & 64.7 (2.5) & - \\ 
\midrule RTMRBM \citep{hu2016temporal} & -& 64.6 \\
\midrule   Deep Autoencoder + SVM \citep{ngiam2011multimodal} & 64.4 (2.4) & - \\ 
\midrule LBP-TOP + SVM \citep{Zhao2009} & - & 58.9 \\
\midrule DCT + DBNF \citep{petridis2016deep} &-& 58.1 \\
\midrule CNN + LSTM \citep{feng2017audio} & 57.7 (0.8) & - \\
\midrule Multiscale Spatial Analysis \citep{matthews2002extraction} & - & 44.6 \\
\bottomrule
\vspace{-0.5cm}
\end{tabularx} 
\end{table}

\begin{table}[t!]
\renewcommand{\tabcolsep}{7pt}
\caption{Classification Accuracy on the AVLETTERS2 database. The end-to-end models are trained using the speaker-independent evaluation protocol \citep{cox2008challenge} where a 5-fold cross-validation is used. ``Mean (Std)'' refers to the mean classification accuracy over ten runs and the corresponding standard deviation, while ``Max'' reports the maximum classification accuracy. RTMRBM: Recurrent Temporal Multimodal Restricted Boltzman Machine, LBP-TOP: Local Binary Patterns-Three Orthogonal Planes, KSRC: Kernel Sparse Representation Classifier. }
\label{tab:resultsAVLETTERS2}
\centering
\begin{tabularx}{\columnwidth}{bsy}
\toprule  Method & Mean (Std)  & Max \\
\midrule End-to-End (Raw Image)   & \textbf{36.8 (2.9)}	& \textbf{42.6}  \\ 
\midrule End-to-End (Diff Image)  &28.9 (2.0) &32.2  \\ 
\midrule End-to-End (Raw + Diff Images)  &35.0 (1.6) &37.8 \\
\midrule RTMRBM \citep{hu2016temporal} & -& 31.2 \\
\midrule LBP-TOP + KSRC \citep{frisky2015lip}  & -& 25.9 \\
\midrule AAM + HMM \citep{cox2008challenge} & - & 8.3 \\
\bottomrule
\vspace{-0.5cm}
\end{tabularx} 
\end{table}

\section{Experimental Setup}

\subsection{Evaluation Protocol}

First, all datasets are divided into into training, validation and test sets. The standard evaluation protocol for the OuluVS2 database is followed  where 40 subjects are used for training and validation and 12 for testing \citep{ouluVS2Protocol}. Then the 40 subjects are randomly divided into 35 and 5 subjects
for training and validation purposes, respectively. This means that there are 1050, 150 and 360 training, validation and test utterances, respectively.

For experiments on the CUAVE database the evaluation protocol suggested in \citep{ngiam2011multimodal} was used. The odd-numbered subjects (18 in total) are used for testing and the even-numbered subjects are used for training. The latter are further divided into 12 subjects for training and 6 for validation.
This means that there are 590, 300 and 900 training, validation and test utterances, respectively.

 The same protocol as the one used in \citep{ngiam2011multimodal}, \citep{matthews2002extraction} is followed for the AVLetters datasets.
 The first two utterances of each subject are used
for training and the last utterance is used for testing. This
means that there are 520 training utterances and 260 test utterances.


The speaker-independent protocol suggested in \citep{cox2008challenge} is used for the AVLetters2 dataset. A 5-fold cross-validation is used, where three speakers are used for training, one for validation and one for testing. This means that in each iteration of the cross-validation there are 546, 182 and 182 training, validation and test utterances, respectively.

\looseness - 1
The target classes are a one-hot encoding for the 10 (case of CUAVE and OuluVS2) or 26 utterances (case of AVLetters and AVLetters2). The label of each utterance is used to label each frame  and the end-to-end model is trained with these frame labels. The majority label over each utterance is used for labeling the entire sequence.

Every time a deep network is trained the results vary due to random initialisation. Hence, in order to present a more objective evaluation  each experiment is repeated 10 times and the mean and standard deviation of classification accuracy on the utterance level are reported.

\subsection{Preprocessing}

The impact of subject dependent characteristics first needs to be reduced since almost all the experiments are subject independent\footnote{Only the evaluation protocol on AVLetters is subject dependent.}.  This is achieved by subtracting the mean image, computed over the entire utterance, from each frame.

\looseness - 1
The next step is the normalisation of data. All images are  z-normalised, i.e. the mean and standard deviation should be equal to 0 and 1 respectively, as suggested in in \citep{hinton2012practical} before pre-training the encoding layers.

\section{Results}
\label{sec:results}

\looseness - 1
In this section we present results for the two-stream end-to-end model, shown in Fig. \ref{fig:system}, and also for each individual stream separately.
We report the mean classification accuracy and standard deviation of the 10 models
trained on each database, OuluVS2, CUAVE, AVLetters and AVLetters2 in Tables \ref{tab:resultsOuluVS} to \ref{tab:resultsAVLETTERS2}, respectively. Just a single accuracy value (with no standard deviation), which is most likely the maximum performance achieved, is provided in almost all previous works. Hence, in order to facilitate a fair comparison, the maximum performance achieved over the 10 runs is also reported.

\looseness - 1
Results for the OuluVS2 database are shown in Table \ref{tab:resultsOuluVS}. The best overall result is achieved by the end-to-end 2-stream model, with a mean classification accuracy of 93.6\%. It is obvious that even the mean performance is consistently higher than the maximum performance of most previous works. When it comes to maximum performance the proposed end-to-end architecture sets the new state-of-the-art on OuluVS2 with 95.6\%. We should also point out, that the proposed 2-stream model outperforms even the CNN models \citep{chung2016lip,chung2016out} trained with external data. Both models are pre-trained on a large dataset, LRW \citep{chung2016lip}, and fine-tuned on OuluVS2. In addition, the proposed model also outperforms the CNN model \citep{han2017multi} trained on all views in a multitask scenario where the goal is to correctly predict both the phrase and the view of the given sequence.

Results for the CUAVE database are shown in Table \ref{tab:resultsCUAVE}. Comparison between different works is difficult since there is not a standard evaluation protocol for this database. The evaluation protocol followed in this study is only used by \citep{ngiam2011multimodal} and \citep{Srivastava:2014}. The best overall performance is achieved by the end-to-end 2-stream model, with a mean classification accuracy of 87.3\% which is an absolute improvement of 18.3\% over \citep{Srivastava:2014}. The maximum classification accuracy of 88.4\% achieved by this models is the new state-of-the-art performance on the CUAVE dataset, which is an absolute improvement of 3.4\% over  \citep{benhaim2013designing}.

Results for the AVLetters database are shown in Table \ref{tab:resultsAVLETTERS}.  The best overall performance is achieved by the end-to-end 2-stream model, with a mean classification accuracy of 66.3\% which is an absolute improvement of 1.6\% over the previous state-of-the-art model \citep{Srivastava:2014}. However, we should note that in this case the improvement over the single stream which uses raw images as input is not statistically significant. The two-stream end-to-end model sets also the new state-of-the-art for the maximum classification accuracy with 69.2\%, which is an absolute improvement of 3.9\% over  \citep{Bakry_2013_CVPR}. At this point we should mention the work of Pei et al. \citep{pei2013unsupervised} which reports a maximum performance of 69.6\%. However, this work uses a non-standard evaluation protocol where the data are randomly divided into 60\% and 40\% for training and testing, respectively.

\begin{figure}[t]
  \centering
\includegraphics[width=0.8\linewidth]{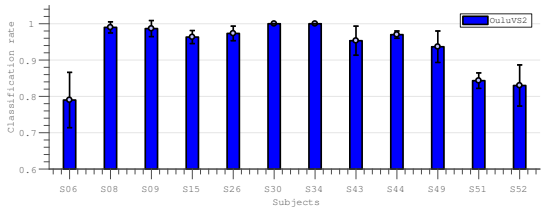}
  \caption{Per-subject performance on OuluVS2 database.}
\label{fig:ouluPerSub}
\end{figure}


\begin{figure}[t]
  \centering
\includegraphics[width=0.8\linewidth]{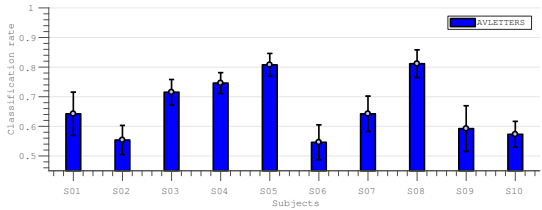}
  \caption{Per-subject performance on AVLETTERS database.}
\label{fig:avlettersPerSub}
\end{figure}


Results for the AVLetters2 database are shown in Table \ref{tab:resultsAVLETTERS2}. In this case, the best overall performance is achieved by the end-to-end single stream model based on raw images, with a mean classification accuracy of 36.8\%.  The main reason the 2-stream does not perform so well is bad tracking of facial points for some subjects. As a consequence, the mouth ROIs extracted are jittery which affects the performance of the \emph{diff} stream. The single stream end-to-end model sets also the new state-of-the-art for the maximum classification accuracy with 42.6\%, which is an absolute improvement of 10.4\% over  \citep{hu2016temporal}. We should emphasize that we use a subject-independent evaluation and due to the small number of subjects the classification accuracy is much lower than the other databases. Much higher results have been reported in the literature for a subject-dependent evaluation protocol with the highest performance of 91.2\% reported in \citep{pei2013unsupervised}.

Fig. \ref{fig:ouluPerSub} shows the classification accuracy per subject for the OuluVS2 dataset.  It is clear that the deviation across  different test subjects  is not very large. Almost all subjects  achieve a classification accuracy over 80\% with 8 of them achieving over 95\%. A similar pattern is also observed in the CUAVE dataset (Figure is not shown due to lack of space).


Fig. \ref{fig:avlettersPerSub} shows the classification accuracy per subject for AVLetters. Contrary to OuluVS2 and CUAVE the performance varies a lot between different subjects with minimum and maximum  accuracies of 54\% and 81\% for subject S06 and S08, respectively. This could be the consequence of the small size of the dataset which does not allow for good generalisation across all subjects or due to differences in the cropped mouth regions. Since the cropped regions are provided it is not easy to verify that all regions were cropped consistently. The same observation about performance variance can be made also for the AVLetters2 dataset (Figure not shown due to lack of space) with a minimum and maximum accuracy of 26\% and 50\% for subjects S05 and S02, respectively. 



The most common confusion pair\footnote{Confusions matrices are not included due to lack of space.} for the OuluVS2 dataset 
is between ``Hello'' (3rd phrase) and ``Thank you'' (8th phrase) which is consistent with confusions presented in \citep{petridis2017deepVisualSpeech,lee2016multi}.
The most frequently confused pairs in the CUAVE dataset 
are  zero and two, and six
and nine  and this is consistent with \citep{petridis2017deepVisualSpeech}. 





The most common confusions for the AVLetters dataset are between B and P, D and T, and U and Q. This is not
surprising since both letters in each pair have the same visual
representation. They consist of two phonemes where the first
ones belong to the same viseme class and the second one
is the same. The letters which are classified correctly most of
the time are the following: M, O, R, W, Y. Similar confusions are observed
on AVLetters2 as well.





Finally, we should also mention that we experimented with CNNs for the encoders but this led to worse performance than the proposed model. This is consistent with the previous results based on CNN models reported on the OuluVS2 and AVLetters databases which are much lower than the proposed system (see the works of \citep{fung2018end, saitoh2016concatenated, lee2016multi} in Table \ref{tab:resultsOuluVS} and \citep{feng2017audio} in Table \ref{tab:resultsAVLETTERS}). This is also reported in \citep{wand2016lipreading} and it is likely due to the small training sets. Only works which have used external data like \citep{chung2016lip,chung2016out} or used all views \citep{han2017multi} have been able to report results based on CNN models on OuluVS2 close to the results presented in this work.

\begin{figure}[t]
    \centering
    \begin{subfigure}[b]{0.48\linewidth}
        \includegraphics[width=\linewidth]{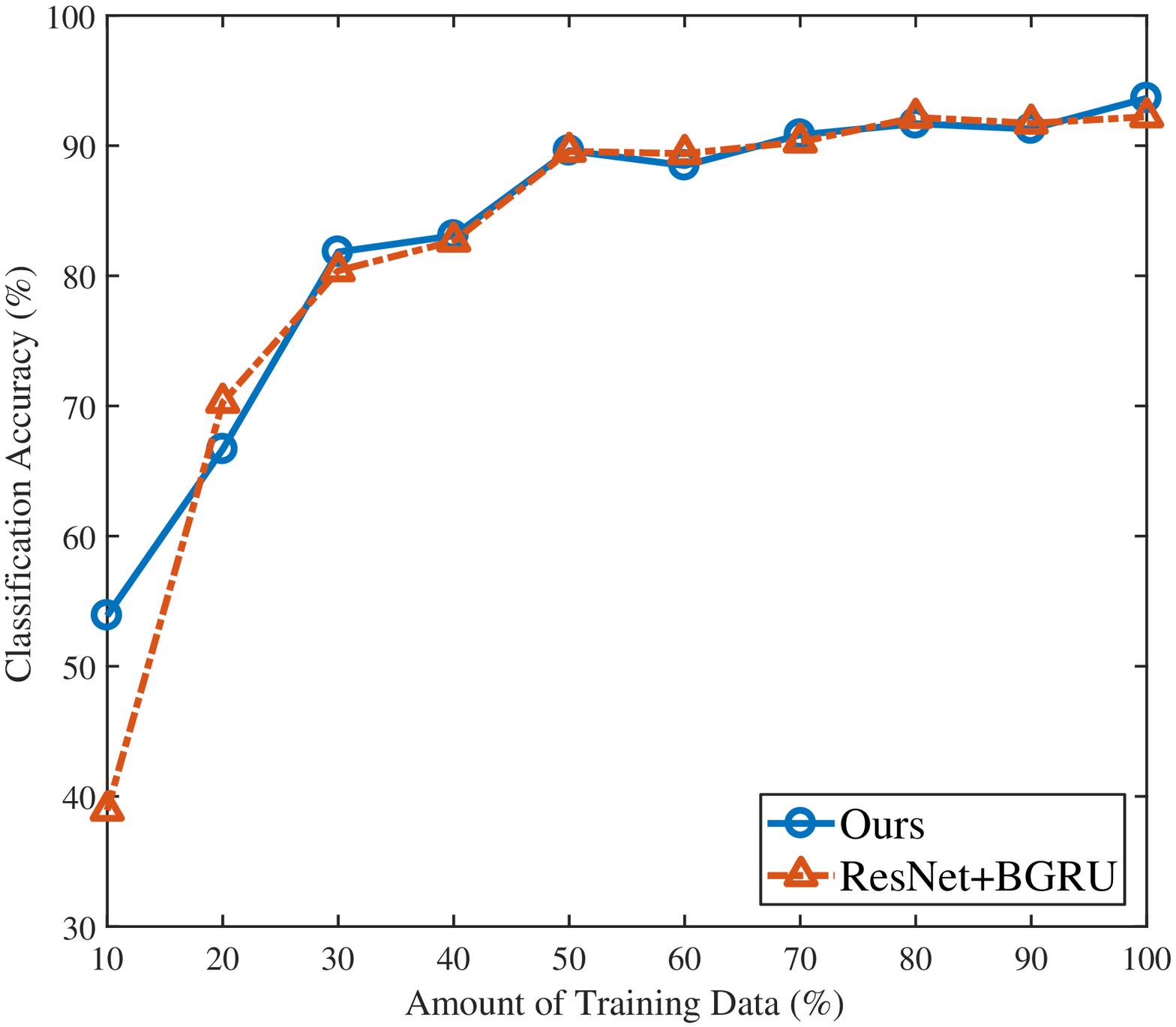}
        \caption{OuluVS2}
        \label{fig:subsetOulu}
    \end{subfigure}
    \begin{subfigure}[b]{0.48\linewidth}
        \includegraphics[width=\linewidth]{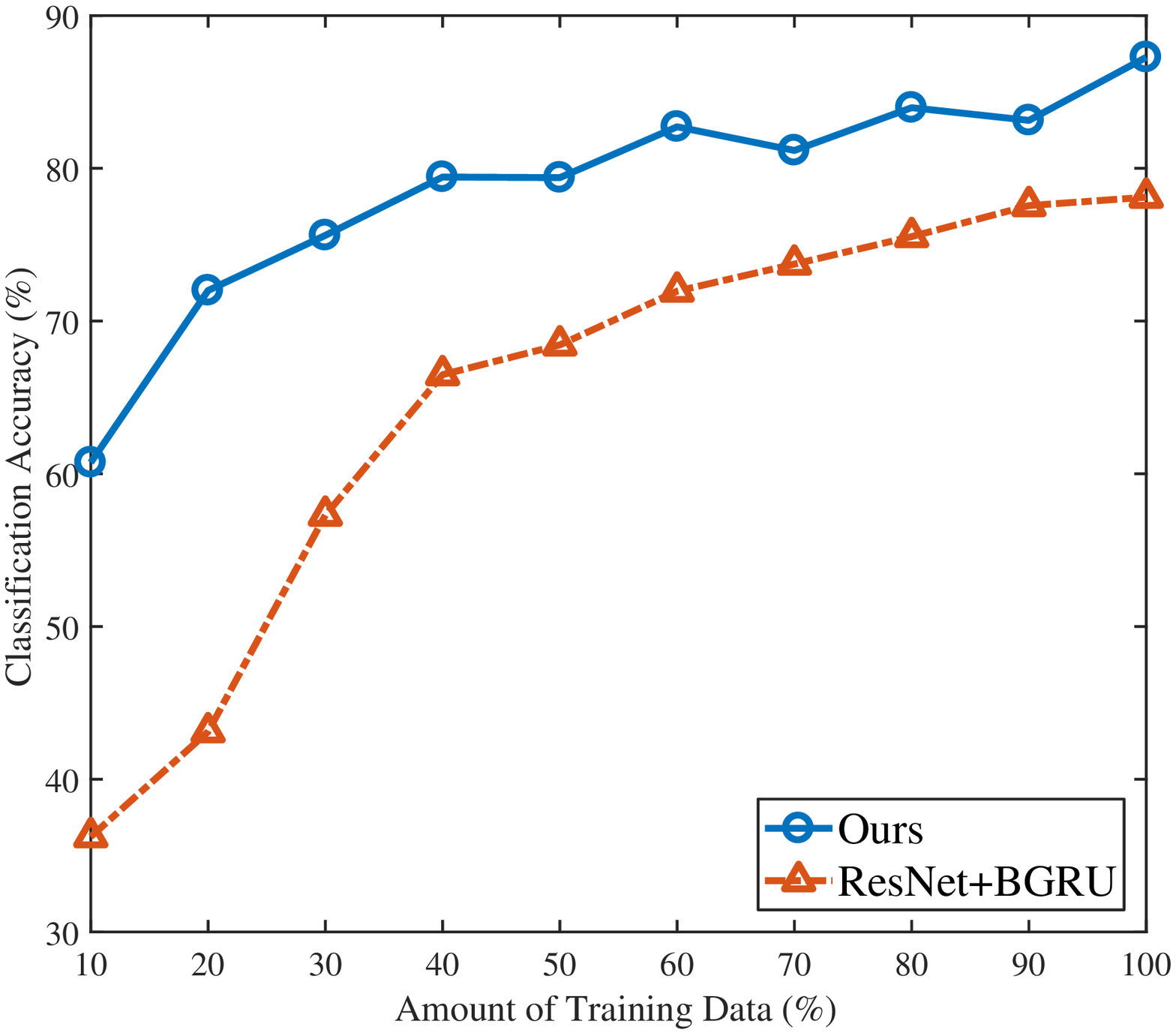}
        \caption{CUAVE}
        \label{fig:subsetCuave}
    \end{subfigure}
    
    \caption{The performance of our approach and the state-of-the-art model based on ResNets and BGRUs \citep{petridis2018AVspeech} as a function of the size of the training set.}\label{fig:animals}
\end{figure}

In order to further test this assumption, we compare the performance of the end-to-end two-stream model with a state-of-the-art lip-reading model as a function of the amount of training data. The model we consider is based on ResNet and BGRUs \citep{petridis2018AVspeech, stafylakis2017combining} and achieves the state-of-the-art performance on the LRW database. The model is trained using the same training protocol as in \citep{petridis2018AVspeech}. Fig. \ref{fig:subsetOulu} and \ref{fig:subsetCuave} show the classification accuracy of the two models for varying training set sizes, from 10\% to 100\%, on the OuluVS2 and CUAVE datasets, respectively. In the former case, the ResNet model quickly reaches the same level of performance as the proposed end-to-end model. In the latter case, the performance gap between the ResNet model and the proposed model decreases as the training set size increases. However, even when the entire training set is used the performance remains below the proposed model. This probably happens due to the small size of CUAVE training set, which is about half the size of the OuluVS2 training set. This is also another indication that CNN models do not reach their full potential for lip-reading applications when trained on small scale datasets and alternative models, like the one proposed here, can be better suited in this scenario.

\vspace{-0.4cm}





\section{Conclusion}
\looseness - 1
In this work, we present an end-to-end visual speech recognition system suitable for small-scale datasets which 
jointly learns to extract features directly from the pixels and perform classification using LSTM networks. Results on four datasets, OuluVS2, CUAVE, AVLetters and AVLetters2, demonstrate that the proposed model achieves state-of-the-art performance on all of them significantly outperforming all other approaches reported in the literature, even CNNs pre-trained on external databases. A natural next step would be to extend the system in order to be able to recognise sentences instead
of isolated words. 


\bibliographystyle{model2-names}
\bibliography{references}

\end{document}